\documentclass{article}

\usepackage{arxiv}

\usepackage[utf8]{inputenc} 
\usepackage[T1]{fontenc}    
\usepackage[bookmarksnumbered=true]{hyperref} 
\hypersetup{
     colorlinks = true,
     linkcolor = blue,
     anchorcolor = blue,
     citecolor = blue,
     filecolor = blue,
     urlcolor = blue
     }
\usepackage{url}            
\usepackage{booktabs}       
\usepackage{amsfonts}       
\usepackage{nicefrac}       
\usepackage{microtype}      
\usepackage{lipsum}
\usepackage{lscape}
\usepackage{graphicx}
\usepackage{lineno}
\usepackage{array}
\usepackage{longtable}
\usepackage{amsmath}
\usepackage{algorithm,algpseudocode}

\title{Optimal DG allocation and sizing in power system networks using swarm-based algorithms}

\author{
  Kayode E. Adetunji \\
  School of Electrical and Information Engineering\\
  University of the Witwatersrand\\
  Johannesburg\\
  \texttt{kayvins@gmail.com} \\
   \And
  Ivan Hofsajer \\
  School of Electrical and Information Engineering\\
  University of the Witwatersrand\\
  Johannesburg\\
  \texttt{ivan.hofsajer@wits.ac.za} \\
     \And
  Ling Cheng \\
  School of Electrical and Information Engineering\\
  University of the Witwatersrand\\
  Johannesburg\\
  \texttt{ling.cheng@wits.ac.za} \\
}

\begin{document}
\maketitle

\begin{abstract}
Distributed generation (DG) units are power generating plants that are very important to the architecture of present power system networks. The benefit of the addition of these DG units is to increase the power supply to a network. However, the installation of these DG units can cause an adverse effect if not properly allocated and/or sized. Therefore, there is a need to optimally allocate and size them to avoid cases such as voltage instability and expensive investment costs. In this paper, two swarm-based meta-heuristic algorithms, particle swarm optimization (PSO) and whale optimization algorithm (WOA) were developed to solve optimal placement and sizing of DG units in the quest for transmission network planning. A supportive technique, loss sensitivity factors (LSF) was used to identify potential buses for optimal location of DG units. The feasibility of the algorithms was confirmed on two IEEE bus test systems (14- and 30-bus). Comparison results showed that both algorithms produce good solutions and they outperform each other in different metrics. The WOA real power loss reduction considering techno-economic factors in the IEEE 14-bus and 30-bus test system are 6.14 MW and 10.77 MW, compared to the PSOs' 6.47 MW and 11.73 MW respectively. The PSO has a more reduced total DG unit size in both bus systems with 133.45 MW and 82.44 MW compared to WOAs'  152.21 MW and 82.44 MW respectively. The paper unveils the strengths and weaknesses of the PSO and the WOA in the application of optimal sizing of DG units in transmission networks.
\keywords{Swarm intelligence \and Optimization algorithms \and Power system networks \and Real and reactive power loss minimization \and Voltage profile improvement \and Economic constraints}
\end{abstract}

\section{Introduction}\label{section1}
The planning of Power System Networks (PSN) with the integration of Distribution Generation (DG) is an important feature in the modern-day electric utility grid. There are certain parameters to be considered such as quantity and size of DG units, the best location, bus configuration, and even the most suitable DG unit technology to be used \cite{Ehsan2018b}. A common and central problem is the placing and sizing of DG units \cite{Ehsan2018b}. The wrong installation of DG units can have an adverse effect on power flow and voltage stability which will in turn cause an upsurge in line losses \cite{Ackermann2001,Balamurugan2012}, thereby inferring an increase in economic costs \cite{DULAU2015}. It is therefore important to find the optimal location and size of DG units in a PSN to solve certain objectives. The installation of DG units is not a linear problem. This means that the increase in the number of DG units does not directly improve grid performance. 

The most reported objective in literature is the power loss minimization. Several methodologies have been proposed for improving common objective functions such as power loss minimization and voltage profile improvement. Analytical approaches (mathematical methods) and metaheuristic methods. Metaheuristic methods have been more consistent because of the combinatorial nature of the placing and sizing problem. Hence, better computational time and efficiency rate can be achieved with less expensive computation resources. In addition to the solving the complex problem, some techniques have been developed to reduce the complexity of the problem. Examples are power loss index (PLI), voltage stability index (VSI), and most recently loss sensitivity factor (LSF). These techniques have also been researched on, and improved overtime \cite{Karimi2017}. 

Over the last decade, meta-heuristic methods have been developed with or without supplementary techniques for optimal location and sizing of DG units in PSNs. Deghanian et al \cite{Dehghanian2013} presented a Non-dominated sorting genetic algorithm (NSGA-II) for finding optimal location of DG units with objective functions of system reliability improvement, total system cost reduction, and network loss minimization in distribution power system. The authors used the probabilistic load flow to simulate uncertainties in a power system. The algorithm was tested on a IEEE 37-bus test system. 

Prakash and Lakshminarayana \cite{Prakash2016a} presented PSO algorithm to place multiple DG units with the objective of minimizing power loss. Jamian et al \cite{Jamian2015} also presented a rank evolutionary PSO for optimal multiple DG placement for maximum power output. The results was compared other variants of the PSO. 

Moradi and Abedini \cite{Moradi2016} proposed a hybrid algorithm, Intelligent Water Drops (IWD) and GA for optimally sizing and allocating DGs in a microgrid, for solving objectives such as power loss minimization and improvement of voltage stability. Voltage Stability Indices (VSI) was used for the identification of candidate buses. The hybrid algorithm was tested on a 33-bus and 69-bus system, and has a record computational time, which increases linearly with number of DG units.

Kaur at al \cite{Kaur2014} developed a mixed integer nonlinear programming (MINLP) for finding optimal placement and sizing of DG units with an objective function of loss minimization in distribution systems. The deterministic approach was carried out through branch and bound method using sequential programming. Combined loss sensitivity was used to shortlist candidate buses which was followed by the implementation of the algorithm. This method differs from approaches where the algorithm is solely used for optimal sizing while the technique would have catered for optimal placement. Scalfati et al \cite{Scalfati2017} used a similar (but mixed integer linear programming) approach to optimally size energy sources in a DC microgrid, given the objective to minimize total cost of ownership. This was implemented for the planning and management phase of the microgrid.

Rastgou et al \cite{Rastgou2018} implemented an improved harmony serach algorithm with sensitivity analysis for network expansion and planning in a DG-present grid network. Karimi et al \cite{Karimi2017} also worked on optimal planning using NSGA-II. The objectives were maximizing voltage stability, minimizing costs, minimizing voltage deviation, and reducing emission. 

HassanzadehFard and Jalilian \cite{HassanzadehFard2018} developed a PSO algorithm to optimally size and site RE-based DG units, considering total cost minimization, total network power loss minimization, voltage stability improvement, and emission reduction, under varying loads. A 13-bus test radial network was used for evaluation. 

Gampa and Das \cite{Gampa2017} presented a two-stage framework to sequentially find optimal location of DG units using sensitivity analysis and a fuzzy-heuristic to optimally size the DG units. This was done for reconfiguration of feeders considering real power loss reduction and voltage profile improvement. 

Jeddi et al \cite{Jeddi2019} implemented harmony algorithm for planning distribution networks. Profit maximization and reliability benefit were considered. Home-Ortiz et al \cite{Home-Ortiz2019} proposed a mixed integer conic programming (MICP) to optimally find type, size, and location of RE-based DG units. 

Ali et al \cite{Ali2017} used the ant lion optimization (ALO) algorithm alongside loss sensitivity factors (LSF) to optimal allocate and size DG units in a distribution network. This was done considering loading conditions and voltage profile improvement. The algorithm was tested on 33 a- and 69-bus test systems. Reddy et al \cite{Reddy2017a} also used the ALO but alongside index vector technique to optimally place and size DG units. The technique was used to select buses for DG locations. The problem was solved considering power loss minimization. Their algorithm was tested on four IEEE test bus systems. The index vector was also used alongside whale optimization algorithm (WOA) \cite{P.2018} to sequentially place and size DG units, considering power loss minimization.

Rammoorthy and Ramachandran \cite{Ramamoorthy2016} applied the PSO to the optimal sizing of multiple DG units in transmission systems. The algorithm was implemented alongside the LSF technique to reduce power loss and enhance the voltage profile. The IEEE 30-bus system was used for evaluation. Mustaffa et al \cite{Mustaffa2017} also implemented an Adaptive Evolutionary Programming method on the IEEE 30-bus system, to reduce the total real power loss. The algorithm outperforms the PSO on four different scenarios.

To the best of the author's knowledge, researchers have not studied the comparison of the application of the PSO and WOA to the optimal sizing problem in a meshed transmission network. More studies have been carried out on RDNs as compared to transmission networks. Most RDN studies consider a void network. On the other hand, transmission networks are considered as an existing network that comprises of units such as transformers, synchronous machines, and synchronous condensers \cite{Schweppe1970}. This study endeavours to uncover the distinctive difference in performance between the implementation of the PSO and the WOA to the optimal sizing problem. Specifically, this paper implements the Loss Sensitivity Factor (LSF) alongside the WOA and the PSO for the optimal location and sizing of DG units in a meshed transmission network. The objective is to compare the total real and reactive power loss minimization and the voltage profile improvement results from the PSO and the WOA.

The rest of the paper is organized as follows. Section \ref{section2} discusses the modelling of a meshed transmission network where the important parameters and the load flow model are addressed. Section \ref{section3} discusses the addition of DG units to the network. The section further explains the technique for the selection criteria. The objective function for the sizing problem is discussed in Section \ref{section4}, where the technical, economic cost and constraints are formulated. In Section \ref{section5}, the PSO and the WOA are presented as the robust swarm intelligence for the sizing problem. Section \ref{section6} presents the results from the analysis. In doing so, the IEEE 14- and 30-bus was used as the testbed for evaluation. Finally, the conclusion is entailed in Section \ref{section7}.

\section{Mathematical Modelling of Meshed Transmission Network}\label{section2}
\begin{figure}[]
	\centering
		\includegraphics[scale=.46]{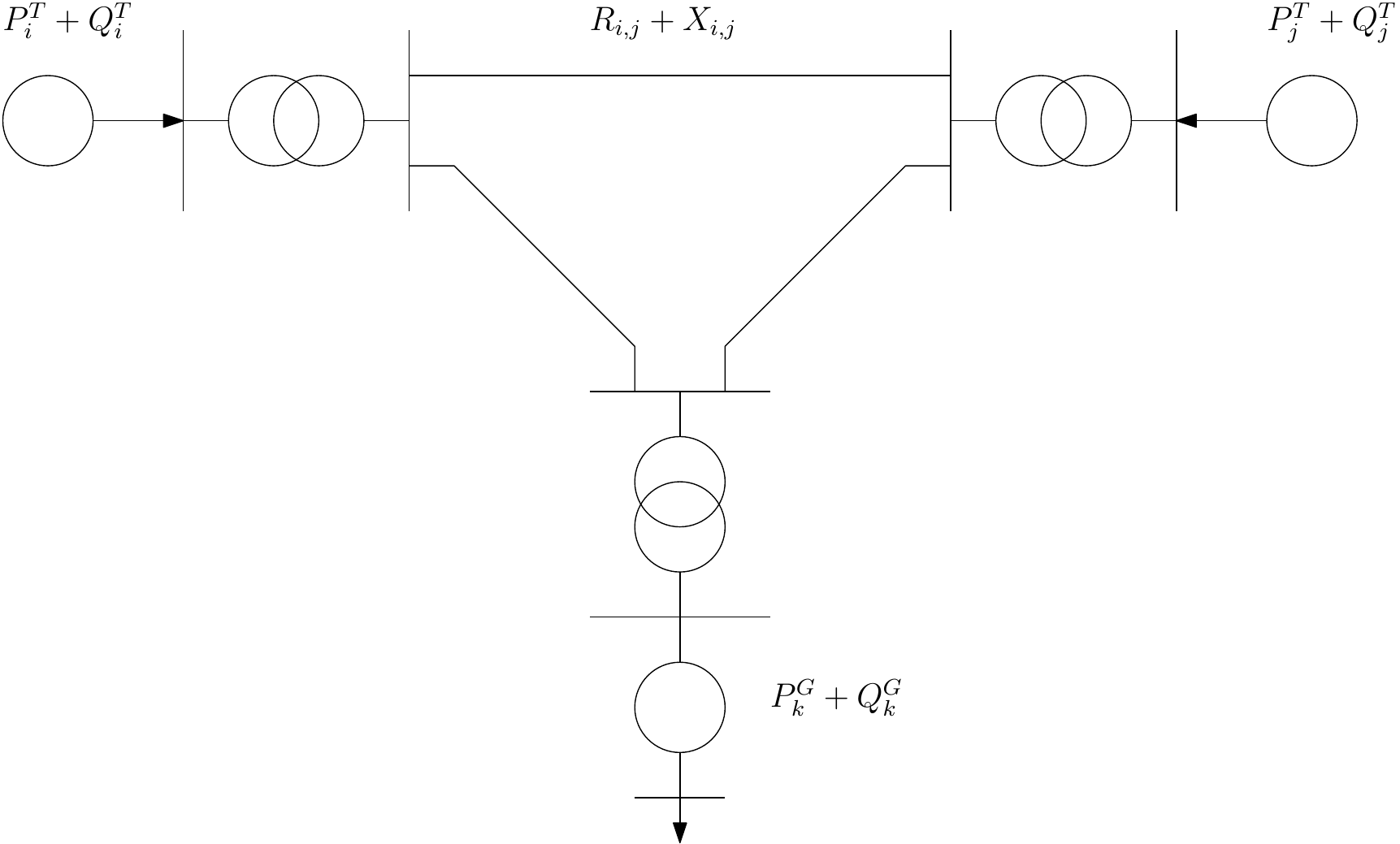}
	\caption{Schematic representation of a simple meshed TN}
	\label{FIG:1}
\end{figure}

The load flow problem is majorly formulated to analyze the normal operation state for power system networks. A practical representation of a meshed Transmission Network (TN) is illustrated in \autoref{FIG:1}. TNs can consist of electrical components such as transformers, capacitors, transmission lines, and impedance loads \cite{Schweppe1970,Kundur1989}. A vital feature for the modelling of a TN is the use of nodes and branches, which represent the buses and transmission lines respectively. Nodes are classified into three major types namely $PQ$, $PV$ and slack nodes. The $PQ$ nodes possess the real and reactive ($P,Q$) parameters and voltage magnitude and the angle will be calculated. The $PV$ nodes are taken as known real power and voltage magnitude, $V$. The slack node is considered as a special bus because of its singularity, i.e. there is only one slack bus in a TN. The known parameters for this node is voltage magnitude and angle ($V,\delta$).

From the diagram, $P_i^T$ and $Q_i^T$ represents an installed substation transformers' real and reactive power from node $i$. $P_k^G$ and $Q_k^G$ represents an installed generators' real and reactive power from node $j$. The impedance on the line is represented as $R_{i,j} + X_{i,j}$. The power equation can be expressed as follows;
\begin{equation}\label{eqn1}
    P_i+Q_i\;=\;\dot V\sum_{j\neq i}{Y}_{ij}{ V}_j\;(i\;=\;1,\;2\;,...\;,\;n),
\end{equation}
where $P_i$ and $Q_i$ are the injected real and reactive power at node $i$ respectively. Here, $ {\widehat Y}_{ij} $ is an element of an admittance matrix from node $ i $ to $j$ and ${\widehat V}_j$ is the injected voltage at node $j$. 

The nodal equation can be written in terms of injected current $I_i$
\begin{equation}\label{eqn2}
    I_i\;=\;\sum_{j\neq i}{Y}_{ij}{V}_j\;(i\;=\;1,\;2\;,...\;,\;n).
\end{equation}
\subsection{Load Flow Studies}
This study implemented the Newton-Raphson (N-R) algorithm to compute the bus voltage magnitudes and angles, real and reactive power of DGs, and real and reactive power loss of the PSN. The N-R algorithm is an iterative method which approximately converts non-linear simultaneous equations to linear simultaneous equations. N-R is known for a fast convergence characteristic, and performs better than most deterministic load flow technique \cite{GopiyaNaik2013}.
For the N-R load flow, the injected current can be represented as shown in Equation \eqref{eqn3},
\begin{equation}\label{eqn3}
    I_i\;=\;\sum_{j\neq i}^N{Y}_{ij}{V}_j\angle (\theta_{ij}+\delta_j).
\end{equation}
The injected current can be substituted for real and reactive power units as
\begin{equation}\label{eqn4}
    P_i-Q_i\;=\; \sum_{j\neq i}^{N}{Y}_{ij}{V}_i{V}_j\angle(\theta_{ij}+\delta_j-\delta_i).
\end{equation}
Therefore, real and reactive power can be derived by splitting Equation \eqref{eqn4} as
\begin{equation}\label{eqn5}
    P_i\;=\; \sum_{j\neq i}^{N}{Y}_{ij}{V}_i{V}_j\cos(\theta_{ij}+\delta_j-\delta_i)
\end{equation}
and
\begin{equation}\label{eqn6}
    Q_i\;=\; \sum_{j\neq i}^{N}{Y}_{ij}{V}_i{V}_j\sin(\theta_{ij}+\delta_j-\delta_i).
\end{equation}
The two equations are written in terms of $V$ and $\delta$, and can be easily manipulated to solve for the Jacobian matrix as shown in Equation \eqref{eqn7} \cite{AlAbri2013,Morison1993},
\begin{equation}\label{eqn7}
    \begin{bmatrix}\Delta P\\\Delta Q\end{bmatrix}\;=\;\begin{bmatrix}J_1&J_3\\J_2&J_4\end{bmatrix}\begin{bmatrix}\Delta\delta\\\Delta V\end{bmatrix},
\end{equation}
where $J_1, J_2, J_3$, and $J_4$ are elements in the Jacobian matrix, and $\Delta P$ and $\Delta Q$ are the respective real and reactive power residuals from the scheduled and calculated values \cite{Morison1993,Gao1992}. They are represented as 
\begin{equation}
    \Delta P_i^k = P_i^{sch} -  P_i^k,
\end{equation} and
\begin{equation}
    \Delta Q_i^k = Q_i^{sch} -  Q_i^k.
\end{equation}
The bus voltages and magnitudes can be estimated as 
\begin{equation}
    V^{k+1} = 
    V_i^{k} +  \Delta V_i^k,
\end{equation} and
\begin{equation}
    \delta^{k+1} = 
    \delta_i^{k} +  \Delta\delta_i^k.
\end{equation}

\section{Addition of DG Units to Power System Networks}\label{section3}
The addition of DG units in a bus network is determined by the characteristics of each bus under normal load conditions. The DG unit supplies real power \cite{ChithraDevi2017} and can be represented as the difference between present real power and real power demand at the same node:
\begin{equation}\label{eqn19}
  P_i= P^{\text{DG}}_i- P^{\text{D}}_i  ,
\end{equation}
where $ P^{\text{D}}_i $  is the load demand at node $ i $ while $ P^{\text{DG}}_i $  is the real power injection from DG unit at node $ i $.

\subsection{Loss Sensitivity Factors}
Loss sensitivity factor technique is an analytical approach to determine a proper location of DG units in a power system network. LSF is one method to lessen the job of an optimization algorithm, hence it is computed to reduce the solutions search space. This method was firstly developed for optimal capacitor placement and is based on the principle of linearization \cite{NitinSinghSmarajitGhosh2015}. Its application for DG placement is not far-off. Sensitivity values are calculated (from base case) for each bus and are ranked in a descending order to form a priority list. The rate of change in real power loss (from Equation \eqref{eqn23}) to the real power injection at a particular bus should be equals zero. The LSF equation \cite{Li2018a,Kanth2014} is shown in \eqref{eqn19}.

Here,
\begin{equation}\label{eqn20}
    \frac{{\delta P}_L}{{\delta P}_j}=2{\sum_{j=1,j\neq i}^{N}{(\alpha_{ij}P_j-\;\beta_{ij}Q_j)}}
\end{equation}
where
\begin{equation}\label{eqn21}
    P_i=\;\frac1{\alpha_k}\left[\beta_kQ_i+{\sum_{j=1,j\neq i}^{N}{(\alpha_{ij}P_j-\;\beta_{ij}Q_j)}}\right].
\end{equation}
Here, $P_i$ is the real power injected at node $i$, which is the difference the generated real power and real potential load demand (from Equation \eqref{eqn19}). Therefore, the real power injected from $i^{th}$ DG units. 
\begin{equation}\label{eqn22}
    P_i^{\text{DG}}= P^\text{D}_i - \;\frac1{\alpha_k}\left[\beta_kQ_i+{\sum_{j=1,j\neq i}^N{(\alpha_{ij}P_j-\;\beta_{ij}Q_j)}}\right].
\end{equation}

The sensitivity values is normalized to values between 0 and 1, using Equation \eqref{eqn27} \cite{El-Fergany2015}.

\begin{equation}\label{eqn27}
    LSF_j = \frac{LSF_j - LSF_{min}}{LSF_{max} - LSF_{min}}\;\;\; j\;\forall \;\{2,\;3,\;.\;.\;.\;,\;N_b\}
\end{equation}

\section{Objective Function Formulation for the Problem}\label{section4}
Optimal DG sizing in RDN reduces the real power loss and improves the voltage profile. The power loss can be represented as \cite{Kansal2013}:
\begin{equation}\label{eqn23}
   P_L=\;{\sum_{i=1}^N{{\sum_{j=1}^N{\left[\alpha_{i,j}(P_iP_j+\;Q_iQ_j)\right]+\;}}}}\beta_{i,j}(Q_iP_j+\;P_iQ_j),
\end{equation}
where
\begin{equation}\label{eqn24}
\alpha_{i,j}=\frac{r_{i,j}}{V_iV_j}{{cos}{\left(\delta_i-\;\delta_j\right)}}
\end{equation}
and
\begin{equation}\label{eqn25}
\beta_{i,j}=\frac{r_{i,j}}{V_iV_j}{{sin}{\left(\delta_i-\;\delta_j\right)}}.
\end{equation}

Here, $i$  and $ j $  are the sending and the receiving bus respectively, while 
 $ Z_{i,j}=\;r_{i,j}+\;{jx}_{i,j} $ represent the branch impedance from bus $ i $ to bus $ j $.
Decrease in real power loss leads to the minimization of DG unit cost. However, increase in DG size may cause an adverse effect on the investment cost. Therefore, there is a need to handle the size of DGs at the potential buses (the location). The addition of DG units can also affect the voltage quality of the network, hence a voltage deviation technique is formulated as
\begin{equation}\label{eqn26}
    VD = \sum_{i=2}^{{N_{PQ}}} |V_i-V_{ref}|,
\end{equation}
to check the voltage profile of each bus. The reference voltage, $V_{ref}$ is set at zero and $V_i$ represents the voltage at each PQ bus after the addition of DG units. The objective functions are subjected to constraints.

\subsection{Economic Objective Function }
The cost generated while sizing the DGs is very important. Therefore, there is a need to minimize it simultaneously with the sizing objective \cite{Al-Foraih2018,Phonrattanasak2010,Elattar2018,Wu2014}.

\begin{equation}\label{eqn28}
    DG_{\text{cost}}=\; \sum_{i=1}^N\sum_{j=1}^{N_{DG}} P_{i}K_{conn}T C^{\text{DG}}P_j^{\text{DG}},
\end{equation}
Where the cost of DG unit is represented as $C^{\text{DG}}$, and $T$ represents the electricity tariff from the DG unit source. Here, $K_{conn}$ is the connection factor to the grid, $P_{i}$ is the injected power at $i^{th}$ number of selected buses for the DG unit placement.  The size of the $j^{th}$ DG unit is represented as $P_j^{\text{DG}}$.

The eco-technical objective function is formulated as the sum of the power loss objective ($P_L$) and the DG cost objective ($DG_{\text{cost}}$).

\subsection{Constraints}
The operating voltage at every bus must satisfy the range at all buses. The bus voltage limit is formulated as
\begin{equation}
    V_i^{min}\leq\;V_i\leq V_i^{max} \; \; i = 1,2,...,N.
\end{equation}

where $ V_i $ is the current voltage at bus $ i $. Then $ V_i^{min} $ and $ V_i^{max} $ are 0.98 and 1.01 respectively.

Rated thermal capacity must not be exceeded at each branch. Therefore, the branch current limit is defined as
\begin{equation}
    |I_{i,i+1}|\leq  I_{i,i+1}^{rated}
\end{equation}
where $i$ and $i+1 $ represent the sending and receiving nodes respectively.
\subsubsection{Power Flow Balance}
The total real and reactive power generation (from all DG units) must be equal to the total real and reactive load and the total real and reactive power loss \cite{Kansal2013}. Therefore, a balance of the power flow is calculated as shown below
\begin{equation}\label{eqn10}
{\sum_{i=1}^{N}{P^{\text{DG}}_i=\;{\sum_{i=1}^{N}{P^{\text{LOSS}}_i}}+\;{\sum_{i=1}^N{P^{\text{LOAD}}_i}}}}
\end{equation}
\begin{equation}\label{eqn11}
{\sum_{i=1}^N{Q^{\text{DG}}_i=\;{\sum_{i=1}^{N}{Q^{\text{LOSS}}_i}}+\;{\sum_{i=1}^N{Q^{\text{LOAD}}_i}}}}
\end{equation}
From Equation \eqref{eqn23},
\begin{equation}
\begin{split}
    \sum({P^{\text{DG}}_i - {P^{\text{LOAD}}_i)}} = \sum_{j=1}^{{N}}V_iV_j[G_{ij}cos(\delta_i - \delta_j) \\
    +B_{ij}sin(\delta_i - \delta_j)]
\end{split}
\end{equation}
\begin{equation}
\begin{aligned}
    \sum({Q^{\text{DG}}_i - {Q^{\text{LOAD}}_i)}} = \sum_{j=1}^{N}V_iV_j[G_{ij}sin(\delta_i - \delta_j) \\
    + B_{ij}cos(\delta_i - \delta_j)]
\end{aligned}
\end{equation}
$ i = 1,2,..., N $\\
where $G_k$ and $B_k$ is the conductance and susceptance of the transmission line $k$ respectively, $V_i$ and $V_j$ are the voltage magnitudes at sending node $i$ and receiving node $j$, $\delta_i$ and $\delta_j$ are the voltage angles wrt $V_i$ and $V_j$.

\section{Proposed Algorithms}\label{section5}

\subsection{Particle Swarm Optimization}
PSO is a computationally intelligent algorithm that uses the behavior of organisms (especially birds and fishes). The synergy and coordination of the behaviour of such organisms is based on the manipulation of distances to reach a desired goal. The organisms are represented as particles and they keep a memory of their coordinates \cite{Jamian2015}. During every iteration, the particles communicate and learn from each other. The best position of each particle is then evaluated and called the personal best (pbest). Then, the overall best position from the personal best of every particle is calculated as the global best (gbest). Every particle moves with respect to a velocity. This velocity is calculated using Equation \eqref{eqn33}.
The PSO is dependent on the number of particles that will be needed for the search of solution space (swarm), weight, cognitive parameter, and social parameter.
The weighting factors of the PSO is very important, as it determines the algorithms' performance. In most cases, the parameters are replicated from previous successful studies. Cognitive, social, and weight are the parameters of the PSO. The random factors are induced for the stochastic nature of the algorithm. Each particles' pbest is calculated as 

\begin{equation}\label{eqn34}
x_{i(t+1)}=\;x_{i(t)}+\;V_{i(t+1)},
\end{equation}
where
\begin{equation}\label{eqn33}
V_{t+1}=wV_{t}+C_1r_1\left(P_{i\left(t\right)}-x_{i\left(t\right)}\right)+C_2r_2(g_{\left(t\right)}-x_{i\left(t\right)}).
\end{equation}

Here, $ x_{i(t)} $ is the position of particle $ i $ in time step $ t $ and $ V_{i(t)} $ is the velocity of particle $ i $ in time step $ t $. The best position of each particles in time step $ t $ is represented as $ P_{i(t)} $ while $  g_{(t)} $ is the global best of all particles in the swarm in time step $t$. $C_1$ and $C_2$ represents the cognitive and social parameter while $r_1$ and $r_2$ represents the induced random parameter. The $ g_{(t)} $ does not have the index $ i $ because it belongs to a whole swarm and not a particle-specific experience. The velocity equation is comprised of components related to swarm behavior. The weight, $ w$ is updated in such a way that it decays with respect to maximum number of iterations. This behavior is illustrated as \cite{Kansal2013a}
\begin{equation}\label{eqn35}
w=\;w_{max}-\frac{w_{max}-w_{min}}{t_{max}}t.
\end{equation}

\begin{figure}[]
	\centering
		\includegraphics[scale=.63]{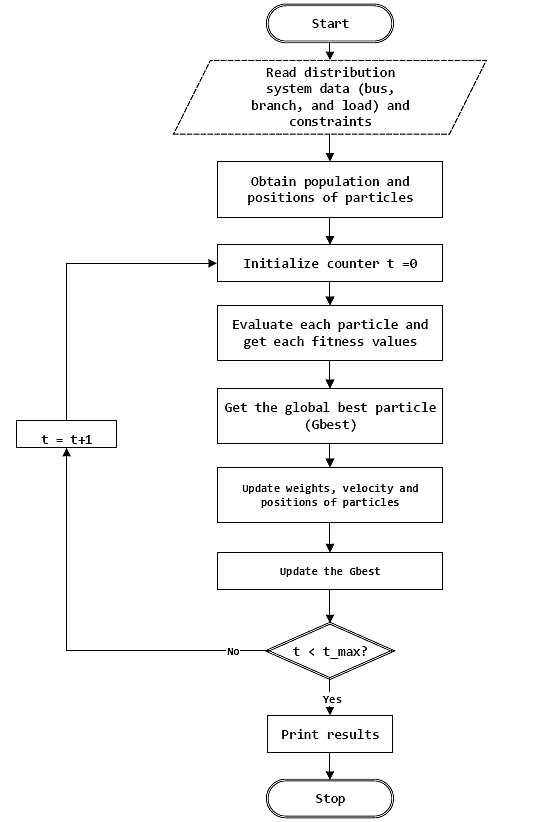}
	\caption{Flowchart of the PSO implementation}
	\label{FIG:2}
\end{figure}

The implementation of the PSO algorithm for optimal sizing is well illustrated in Algorithm \ref{PSOalgorithm}.
\begin{algorithm}
\caption{PSO Implementation}
\label{PSOalgorithm}
Input bus and branch data parameters and run the load flow algorithm to update the bus and branch data.\\ 
Solve the load flow problem to obtain the real and reactive power loss and the voltage (pu) at each bus.\\
Identify the candidate buses by using Equation \eqref{eqn20} to get the sensitivity values. 
\\ Initialize: a population of particles with values and positions from the candidate bus matrix, cognitive and social parameters, number of iterations \textit{maxIter}, and counter \textit{t}.
\begin{algorithmic}

\While{$t < maxIter $}
\For{Each particle $i$}
    \State Adapt velocity $V_n$ of the particle using \eqref{eqn35}
    \State Update the position $x_n$ of the particle using Equation \eqref{eqn36}
    \State Evaluate the objective functions  {$f(\overrightarrow{X}_n)$} - Equation \eqref{eqn23} and \eqref{eqn26}.
    \If{$f(\overrightarrow{X}_n)<f(\overrightarrow{P}_n)$}
       \State $\overrightarrow{P}_i \gets \overrightarrow{X}_i$
    \EndIf
    \If{$f(\overrightarrow{X}_n)<f(\overrightarrow{P}_g)$}
       \State $\overrightarrow{P}_g \gets \overrightarrow{X}_i$
    \EndIf
\EndFor
\EndWhile\\
\Return $\overrightarrow{P}_g$
\end{algorithmic}
\end{algorithm}

\subsection{Whale Optimization Algorithm}
The Whale optimization algorithm (WOA) was proposed by Mirjalili \cite{Mirjalili2016}. The algorithm has been successfully applied to many optimization problems including the optimal sizing problem. The algorithm involves the feeding nature of whales (specifically humpback whales). According to the intelligent behavioural nature of whales, they use a specific technique to target small fish that are close to the sea surface. This technique is called the bubble net feeding, where whales swim round a school of fish (prey) to form a 9-shaped bubble trail. 
In this algorithm, an initial solution is termed as the objective prey and assumed as the current best solution. During the course of iterations, other whales update their positions towards the best whale position.
The WOA is mathematically modelled in three sections. (i) encircling prey (ii) bubble net hunting method and (iii) search the prey.
\subsubsection{Encircling prey}\label{section5.2.1}
This form of hunting is based on a circular positioning around a prey. The position of the whale moves towards the prey in a step-wise manner. It can be modelled as
\begin{equation}\label{eqn36}
\overrightarrow{X}_{t+1}={\overrightarrow X}_{t}^\ast- A\cdot\overrightarrow D,
\end{equation}
where 
$A=2a\;\cdot\;r\;\;-\;a $ and $ a $ is linearly reduced from 2 to 0, and $ r \in [0,1]$ where $ r $ can take up random in $ 0 $ and $ 1 $ interval.\\
$ \overrightarrow D=\;| C\;\;\cdot\;\;\overrightarrow{X_{t}^\ast}-\;\overrightarrow{X_{t}}|, $\\
where 
$ C=2\cdot\;r. $\\
The current iteration is denoted by subscript $ t $ and $ t+1 $ for the next iteration. $ \overrightarrow{X} $ represents the position vector and $ \overrightarrow{X^\ast} $ represents the current best solution. 
\subsubsection{Bubble net hunting method}
\begin{figure}
	\centering
		\includegraphics[scale=.65]{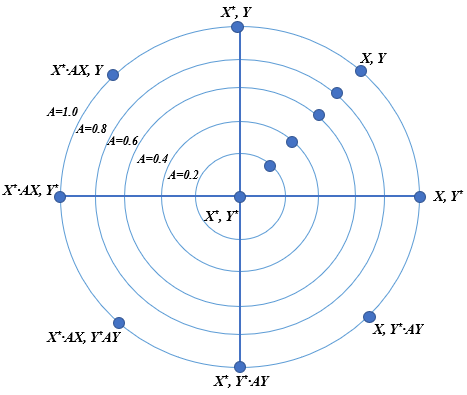}
	\caption{Shrinking mechanism of the WOA (adapted from \cite{Prakash2016a})}
	\label{FIG:3}
\end{figure}

\begin{figure}
	\centering
		\includegraphics[scale=.70]{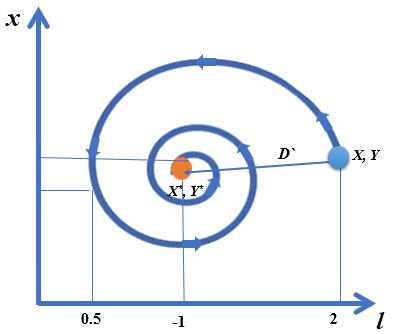}
	\caption{Spiral hunting simulation of the WOA (adapted from \cite{Prakash2016a})}
	\label{FIG:4}
\end{figure}

This method is the exploitative phase of the algorithm. \autoref{FIG:3} illustrates the mechanism of the shrinking approach. This behaviour is attained by decreasing the value of $a$, as explained in Section \ref{section5.2.1}. This behaviour is modelled as shown in Equation \eqref{eqn36}. The spiral hunting approach is a helic-shaped movement of whales from a position $(X, Y)$ to hunt prey (the best solution) at a position $(X*, Y*)$. This is illustrated in \autoref{FIG:4}. The spiral equation is created as follows: 
\begin{equation}\label{eqn37}
\overrightarrow{X}_{t+1}\;=\;\;\overrightarrow{D}'\cdot\;e^{bl}\cdot{{cos}{\left(2\pi l\right)}}+\;\overrightarrow{X}^\ast .
\end{equation}
The component $ cos(2 \pi l) $ simulates the spiral shape of the whale's path, as shown in \autoref{FIG:4}. $ l $ takes a value between [-1,1], and $ b $ is a constant (usually 1) that gives the spiral shape regular definition.
Since the whales can be shrink towards a prey or move spirally, there is a need to model this behaviour. A probability of 50\% is choosen for both feeding mechanisms. The model is as follows:
\begin{equation}\label{eqn38}
\overrightarrow{X}_{t+1}\;=\;\left\{\begin{array}{lc}\; \overrightarrow{X}_t^\ast -\;\overrightarrow{A} \cdot\overrightarrow{D} ,&p<0.5\\\overrightarrow{D} \cdot\;e^{bl}\cdot\cos(2 \pi l)+\;\overrightarrow{X}_t^\ast &p\geq0.5\end{array}\right.
\end{equation}
where $ p $ is a random number between $[0,1]$.
\subsubsection{Search for prey}
The WOA uses this technique to mainly overcome a possible local optimum. This is called the exploration phase, where whales search randomly according to check other possible solutions. By this, if the best solution is not global the search agent can still fall out to find other better solution in the global space. The use of this approach is dependent on the value of $ A $. If the value of $ A $ is lesser than 1, Equation \eqref{eqn36} is triggered, else Equation \eqref{eqn39} is triggered. \autoref{eqn39} shows the model.
\begin{equation}\label{eqn39}
\overrightarrow{X_{t+1}}=\overrightarrow{X_{rand}}- A\cdot\overrightarrow D
\end{equation}
where \\
$ \overrightarrow D=\;\left| C\;\;\cdot\;\overrightarrow{X_{rand}}\;-\;\overrightarrow{X_{t}}\right| $, \\
$ \overrightarrow{X_{rand}} $ represents the random whales' position in current iteration. $ \overrightarrow{D} $ is the distance between current whale position, $ \overrightarrow{X_{t}} $ and the randomly selected whales' position. 
\begin{figure*}[]
	\centering
		\includegraphics[scale=.55]{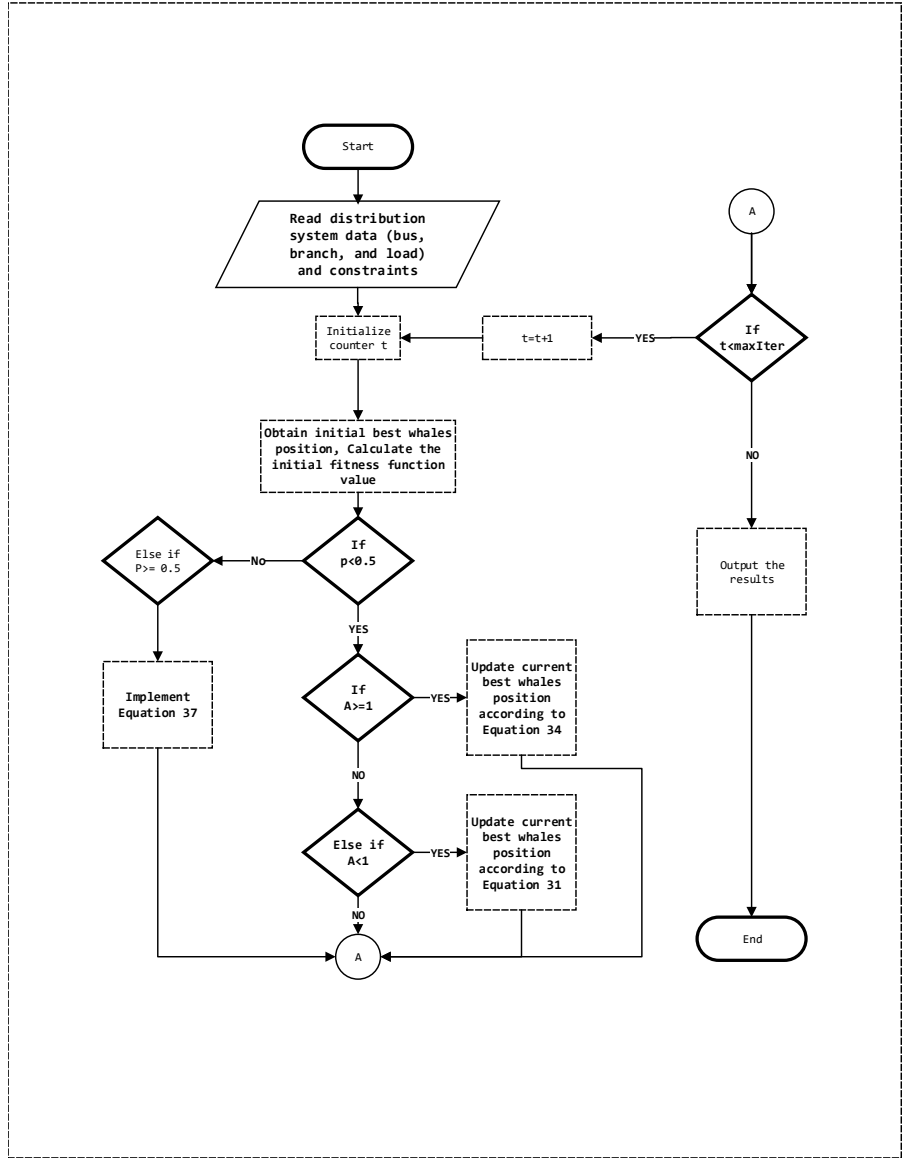}
	\caption{Flowchart of the WOA implementation}
	\label{FIG:5}
\end{figure*}

The implementation of the WOA algorithm for optimal sizing is well illustrated in Algorithm \ref{WOAalgorithm}.

\begin{algorithm}
\caption{WOA Implementation}
\label{WOAalgorithm}
Input bus and branch data parameters and run the load flow algorithm to update the bus and branch data.\\ 
Solve the load flow problem to obtain the real and reactive power loss and the voltage (pu) at each bus.\\
Identify the candidate buses by using Equation \eqref{eqn20} to get the sensitivity values. 
 Initialize: a population of whales with values and positions from the candidate bus matrix, number of iterations \textit{maxIter}, and counter \textit{t}.\\
 Initialize a, A, C, l and p 

\begin{algorithmic}
\State Evaluate the objective functions for each whale  {$f(\overrightarrow{X}_i)$}.
\State Obtain the best whale, $\overrightarrow{X}^*$
\While{$t < maxIter $}
\For{Each particle $i$}
    \If {$p < 0.5$}
        \If{$|A| <1$}
            \State Update the position of the current whale by the equation \eqref{eqn36}
        \Else[$|A| \geq 1$]
                \State Update the position of the current whale by the equation \eqref{eqn39}
        \EndIf
    \Else[$p \geq 0.5$]
        \State Update the position of the current whale by the equation \eqref{eqn37}
    \EndIf
    \State Calculate the fitness of each whale, $\overrightarrow{X}_i$
    \If{$\overrightarrow{X}_i>\overrightarrow{X}^*$}
       \State $\overrightarrow{X}^* \gets \overrightarrow{X}_i$
    \EndIf
\EndFor
\State $t++$
\EndWhile\\
\Return $\overrightarrow{X}^*$
\end{algorithmic}
\end{algorithm}

\section{Implementation Results and Discussion}\label{section6}
The algorithms were developed in the MATLAB environment (R2017a version) and  simulations were run on an Intel(R) Core(TM) i7-3520M @ 2.90 GHz. The outcomes from the algorithms are presented in this section, and have been evaluated on the IEEE 14-bus and 30-bus test RDN system.

Parameters used in the algorithms are presented in \autoref{tbl1}. The metric is based on power loss minimization and voltage profile improvement. The approach uses a two-stage procedure for the optimal location and sizing of DG units. The LSF technique was used to identify the potential buses for placing the DG units. Then the meta-heuristic optimization algorithm was applied to calculate the optimal sizes of the DG units with respect to the potential locations. This two-stage approach is necessary for reducing the solution search space for the algorithms. The main work of the algorithms was done by the N-R load flow algorithm, which mainly uses the bus and branch data to compute the power losses. The parameter values of the algorithms are shown in \autoref{tbl1} \cite{Eberhart2000}.

\begin{figure}
	\centering
		\includegraphics[scale=.75]{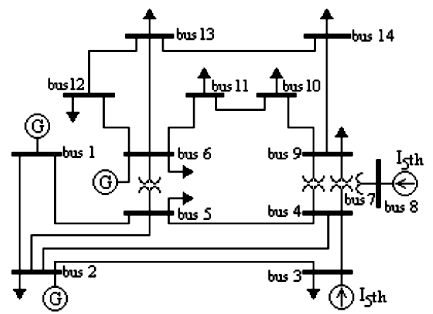}
	\caption{IEEE 14-Bus distribution test system \cite{Allan2010}}
	\label{FIG:6}
\end{figure}

\begin{table}
 \caption{Algorithm parameters}
  \centering
  \begin{tabular}{lll}
    \toprule
    \multicolumn{2}{c}{PSO}                   \\
    \cmidrule(r){1-2}
\textbf{Parameters} & \textbf{Description} & \textbf{Value} \\ 
\midrule
	POP & Population of bird flocks & 50 \\ 
	dim & Dimension & 4 \\ 
	maxIter & Number of iterations & 150 \\ 
    $w_1$ & minimum weight & 0.4 \\
    $w_2$ & maximum weight & 0.9 \\
    $c_1$ & cognitive parameter & 2\\
    $c_2$ & social parameter & 2 \\
\midrule
\multicolumn{2}{c}{WOA}                   \\
    \cmidrule(r){1-2}
	POP & Population of whales & 50 \\ 
	dim & Dimension & 4 \\ 
	maxIter & Number of iterations & 150 \\
\bottomrule
  \end{tabular}
  \label{tbl1}
\end{table}

\subsection{IEEE 14-Bus test system results}
The DG unit sizes were initially chosen randomly within the DG limits $(1 MW\le\ DG size\le\ 50 MW) $ for the potential bus locations. Thereafter, the objective function was computed based on the defined constraints. The total power losses were calculated for each algorithm. The base case was 100 MVA and 11 kV, with voltage magnitude limits between 1.01 and 0.98 pu, with the first bus assumed at 1.0 pu. The total load demand of the test system is 259 MW and 73.5 MVaR, with the real and reactive power loss as 13.593 MW and 56.910 respectively.

\begin{figure}[]
	\centering
		\includegraphics[scale=.6]{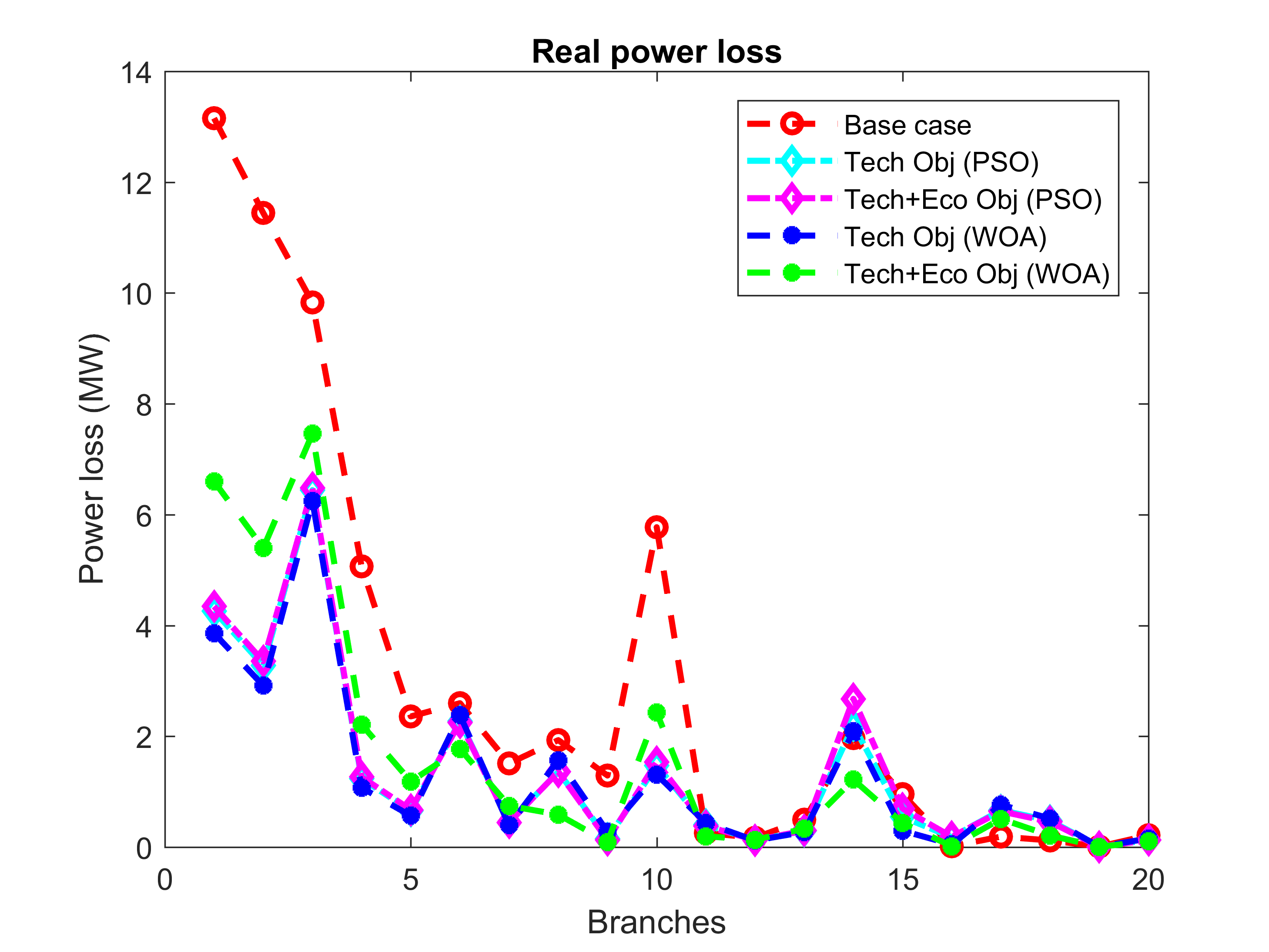}
	\caption{Real power loss on each branch before and after DG integration for 14-bus test system. }
	\label{FIG:7}
\end{figure}
\begin{figure}[]
	\centering
		\includegraphics[scale=.55]{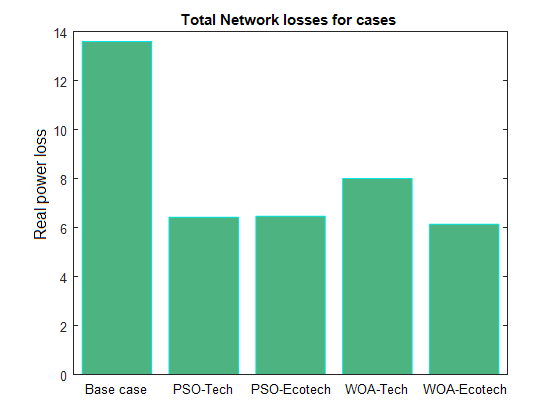}
	\caption{Total real power loss of the IEEE-14 bus system}
	\label{FIG:8}
\end{figure}

\begin{figure}[]
	\centering
		\includegraphics[scale=.55]{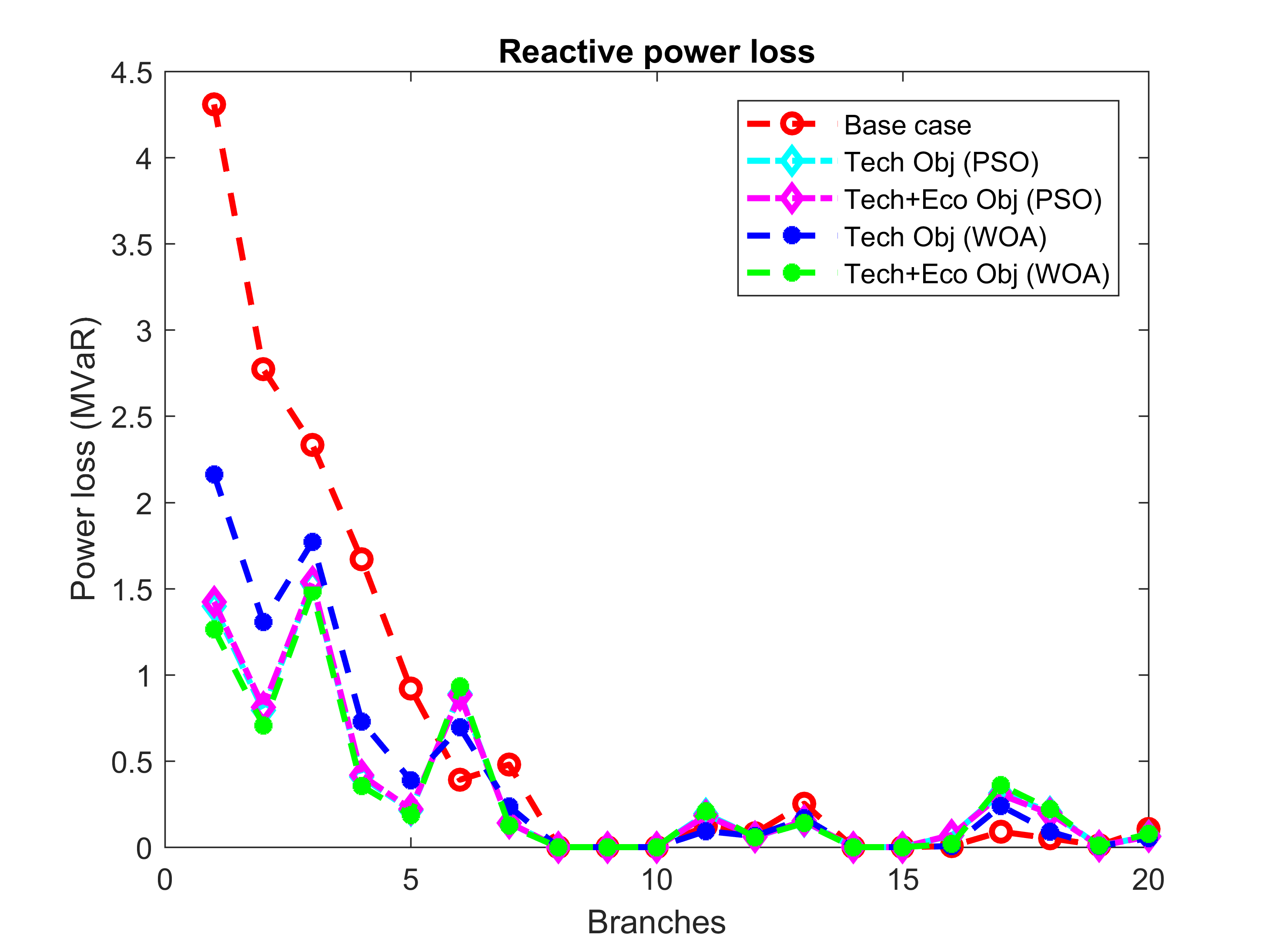}
	\caption{Reactive power loss improvement of IEEE 14-bus test system}
	\label{FIG:9}
\end{figure}

\begin{figure}[]
	\centering
		\includegraphics[scale=.55]{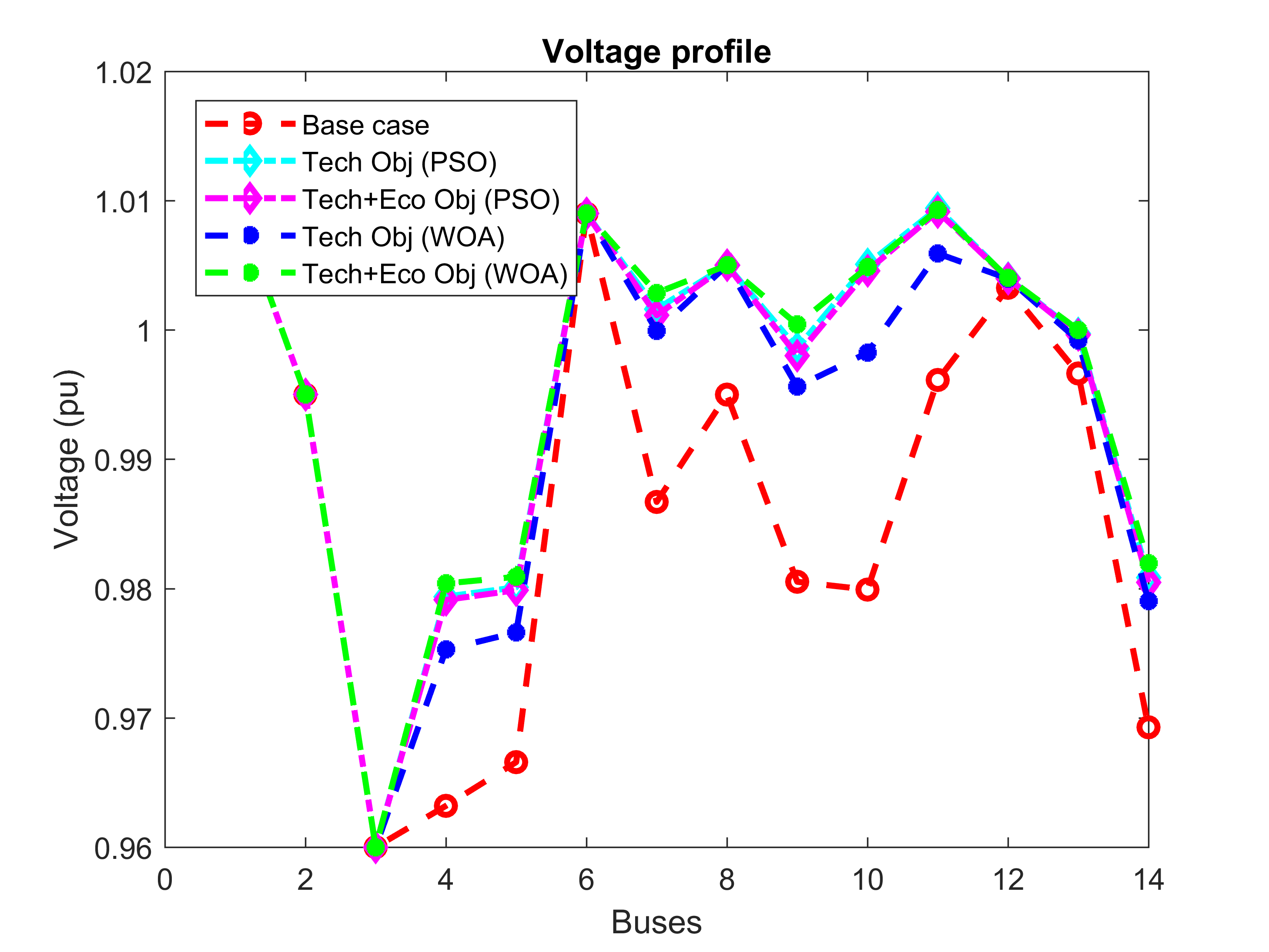}
	\caption{Voltage profile of IEEE 14-bus test system }
	\label{FIG:10}
\end{figure}

\begin{table}
 \caption{Real and reactive power loss values and improvement for 14-bus test system}
  \centering
  \begin{tabular}{lcc}
    \toprule
\textbf{Real power} & \textbf{MW} & \textbf{Performance (\%)} \\
\midrule
WOA Power loss technical & 8.00 & 69.76\% \\
WOA Power loss techno-economic & 6.14 & 121.50\%   \\
PSO Power loss technical & 6.43 & 111.49\% \\
PSO Power loss techno-economic & 6.47 & 110.41\%   \\
\midrule
\textbf{Reactive power} & \textbf{} \\
\midrule
WOA Power loss technical & 27.59 & 07.94\% \\
WOA Power loss techno-economic & 28.44 & 120.58 \%   \\
PSO Power loss technical & 25.80 & 113.65\% \\
PSO Power loss techno-economic & 31.64 & 110.36\%   \\

\bottomrule
  \end{tabular}
  \label{tbl2}
\end{table}

\begin{table}
\caption{Comparison of the algorithms' real and reactive power loss values for 14-bus test system}\label{tbl2a}
\centering
\begin{tabular}{lcc}
\toprule
\textbf{Real power} & \textbf{Performance (\%)} \\
\midrule

WOA improvement technical & -19.73\%   \\
WOA improvement techno-economic & 5.41\%  \\
\midrule
\textbf{Reactive power} & \textbf{} \\
\midrule

WOA improvement technical & -12.73\%   \\
WOA improvement techno-economic & 5.41\%  \\
\bottomrule
\end{tabular}
\end{table}

The LSF was implemented to identify the candidate bus locations for optimal sizing of the DG units. The potential bus locations are 2, 8, 9, 10. These bus locations are selected based on the values close to zero, and are prioritized in a descending manner. The first four indices of the bus configuration are selected. 

\autoref{FIG:7} illustrates the real power loss on each bus. The PSO while considering techno-economic objective, reduces the real power loss. The WOA comes close but only with the consideration of the technical objective. A summary of the total power loss reduction is illustrated in \autoref{FIG:8}, which shows that the PSO averagely performs better than the WOA. \autoref{tbl2} further elucidates the values and improvement level of the algorithms. Each parameter is compared to the base case to calculate the improvement percentage. The PSO, considering the technical and techno-economic objectives, reduced the real power loss to 6.43 MW and 6.47 MW respectively, while the WOA considering the technical and techno-economic objectives, reduced the real power loss to 8.0069 MW and 6.14 MW respectively. The PSO, considering the technical and techno-economic objectives, reduces the reactive power loss to 25.80 MVaR and 31.64 MVaR respectively, while the WOA considering the technical and techno-economic objectives, reduces the reactive power loss to 27.59 MVaR and 28.44 MVaR respectively.  \autoref{tbl3} shows the comparison of the WOA to the PSO in the optimal sizing of DG units. In other words, the PSO is the base case. It is more evident that the PSO performs better than the WOA when considering only the technical objective.

From \autoref{tbl4}, the WOA produces a lower DG unit size when considering only technical objective. However, the PSO performs better than the WOA in terms of the techno-economic objective. This is synonymous to WOA results in \cite{Prakash2017b}. The real and reactive power loss from the PSO is 6.47 MW and 28.43 MVaR, while the WOA is 6.14 MW and 25.80 MVaR respectively. From \autoref{FIG:9}, it is observed that the WOA when considering the techno-economic objective, has the highest reactive power loss reduction on the bus network. \autoref{FIG:10} shows the voltage profile of the IEEE 14-bus test system, with WOA considering techno-economic constraints having the most improved profile.

\begin{table}[h]
\caption{DG units sizes for candidate buses}\label{tbl3}
\centering
\begin{tabular}{lcccc}
\toprule
Bus no. & PSO & PSO (Eco) & WOA & WOA (Eco)\\
\midrule
2 & 41.191 & 39.545 & 20.343 & 49.328 \\
8 & 42.133 & 43.167 & 20.552 & 38.826 \\
9 & 13.823 & 9.880 & 22.645 & 32.311 \\
10 & 38.777 & 37.937 & 20.612 & 33.540 \\
\bottomrule
\textbf{TOTAL} & \textbf{135.924} & \textbf{130.529} & \textbf{84.152} & \textbf{154.005} \\
\bottomrule
\end{tabular}
\end{table}

\subsection{IEEE 30-Bus test system results}
The DG unit sizes were initially chosen randomly within the DG limits $(1\ MW\le\ DG size\le\ 50\ MW) $ for the potential bus locations. Thereafter, the objective function was computed based on defined constraints. The total power losses are calculated for each algorithm. For the base case of 100 MVA and 11 kV, with voltage magnitude limits between 1.01 and 1.1 pu, the real power loss is 13.5929 MW. The total load demand of the test system is 283.4 MW and 126.2 MVaR.

    \begin{figure}[]
	\centering
		\includegraphics[scale=.75]{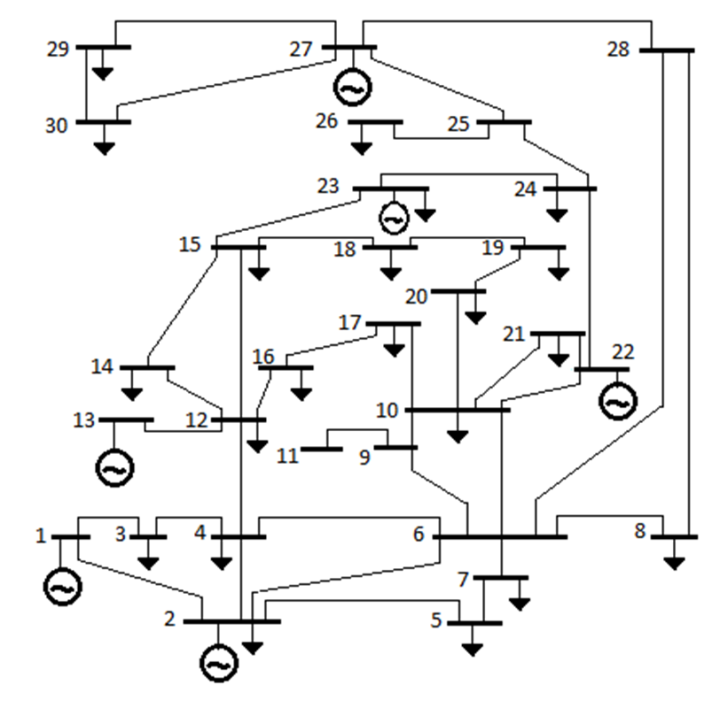}
	\caption{IEEE 30-Bus distribution test system \cite{Hota2016}}
	\label{FIG:11}
\end{figure}
\begin{figure}
	\centering
		\includegraphics[scale=.54]{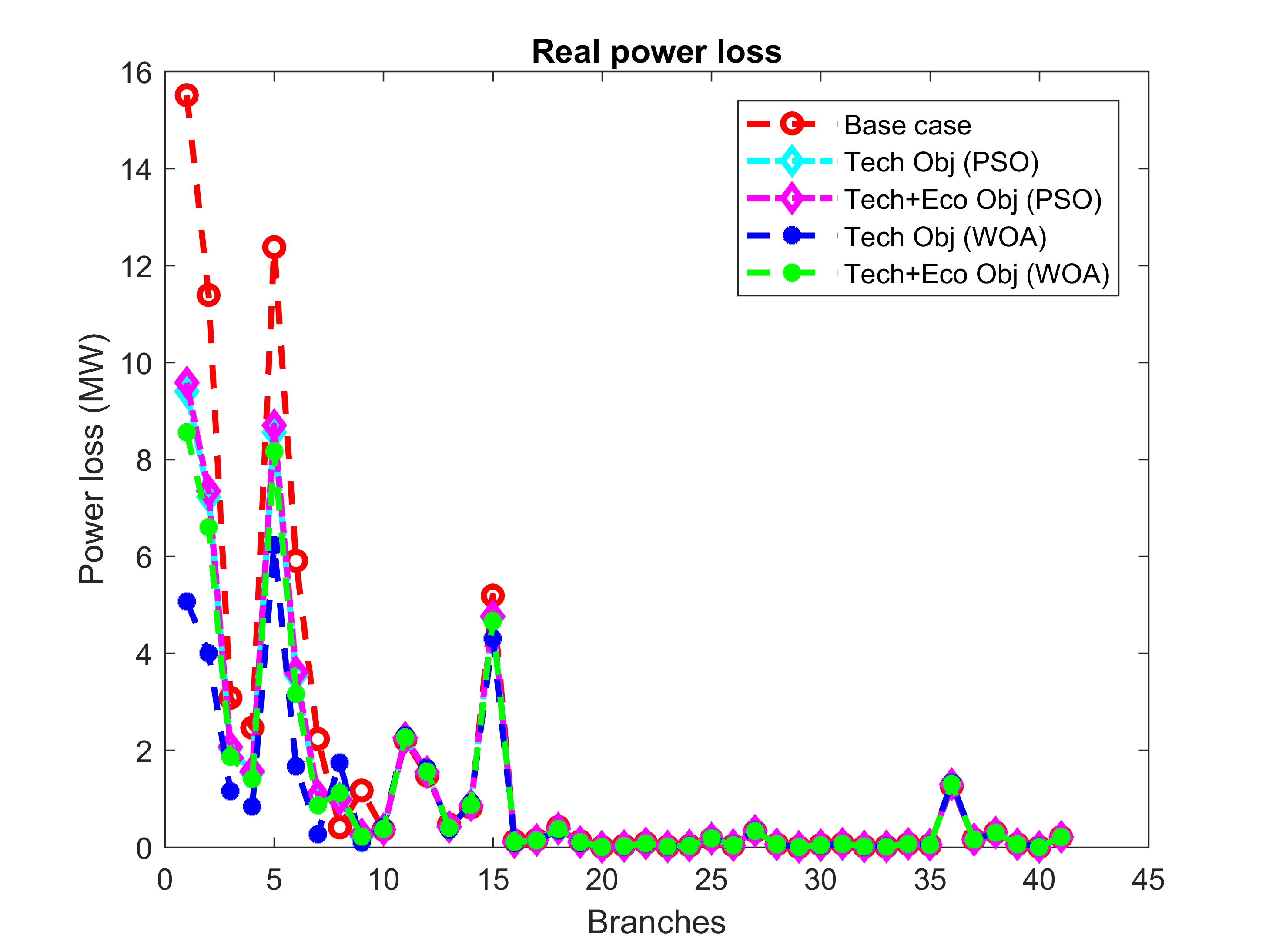}
	\caption{Real power loss on each branch before and  after DG integration for 14-bus test system. }
	\label{FIG:12}
\end{figure}
\begin{figure}
	\centering
		\includegraphics[scale=.55]{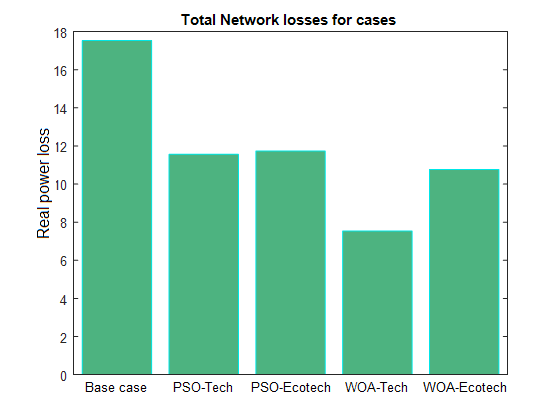}
	\caption{Total real power loss of the IEEE-30 bus system}
	\label{FIG:13}
\end{figure}
\begin{figure}
	\centering
		\includegraphics[scale=.5]{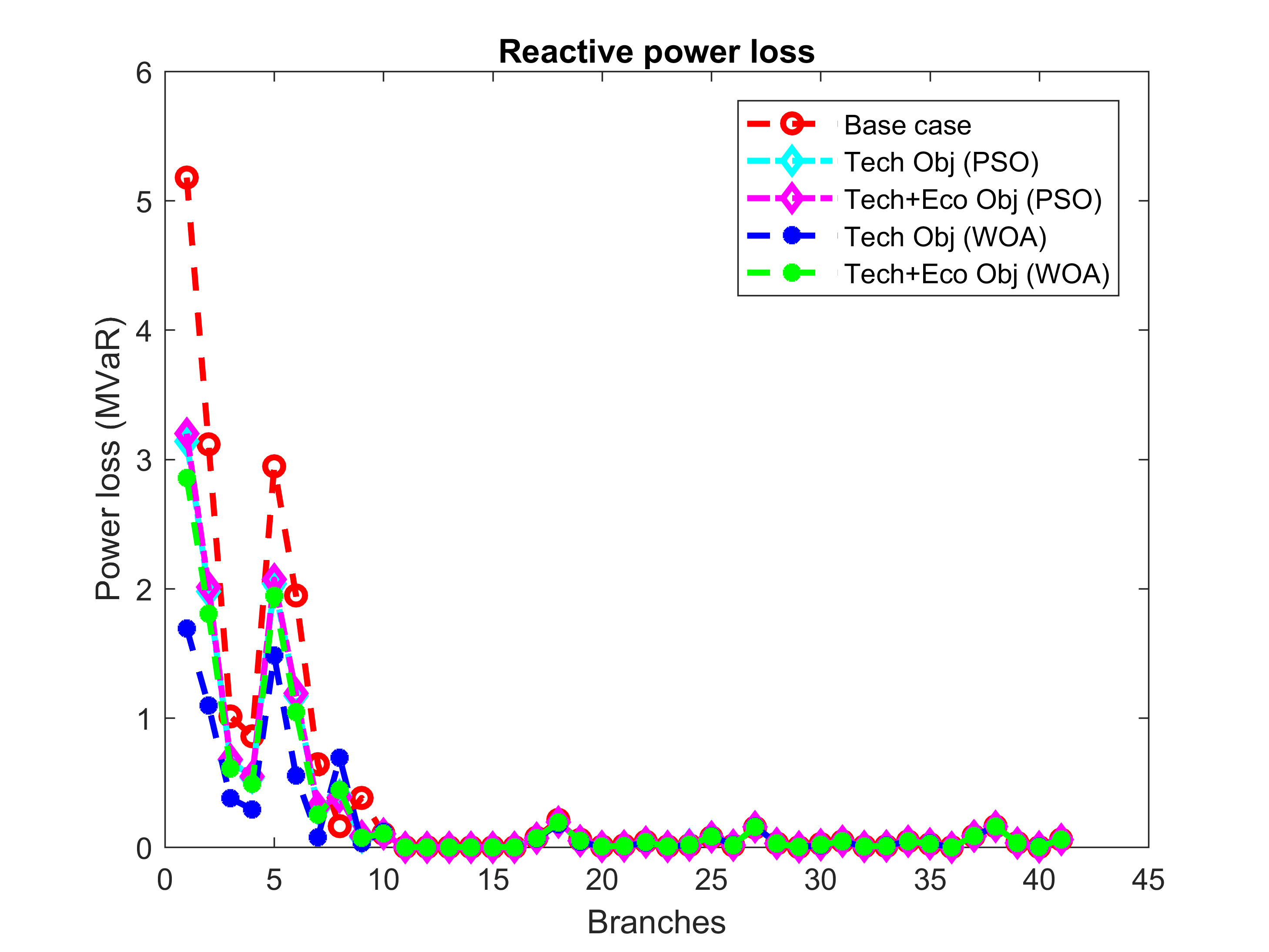}
	\caption{Reactive power loss improvement of the IEEE-30 bus test system}
	\label{FIG:14}
\end{figure}
\begin{figure}
	\centering
		\includegraphics[scale=.5]{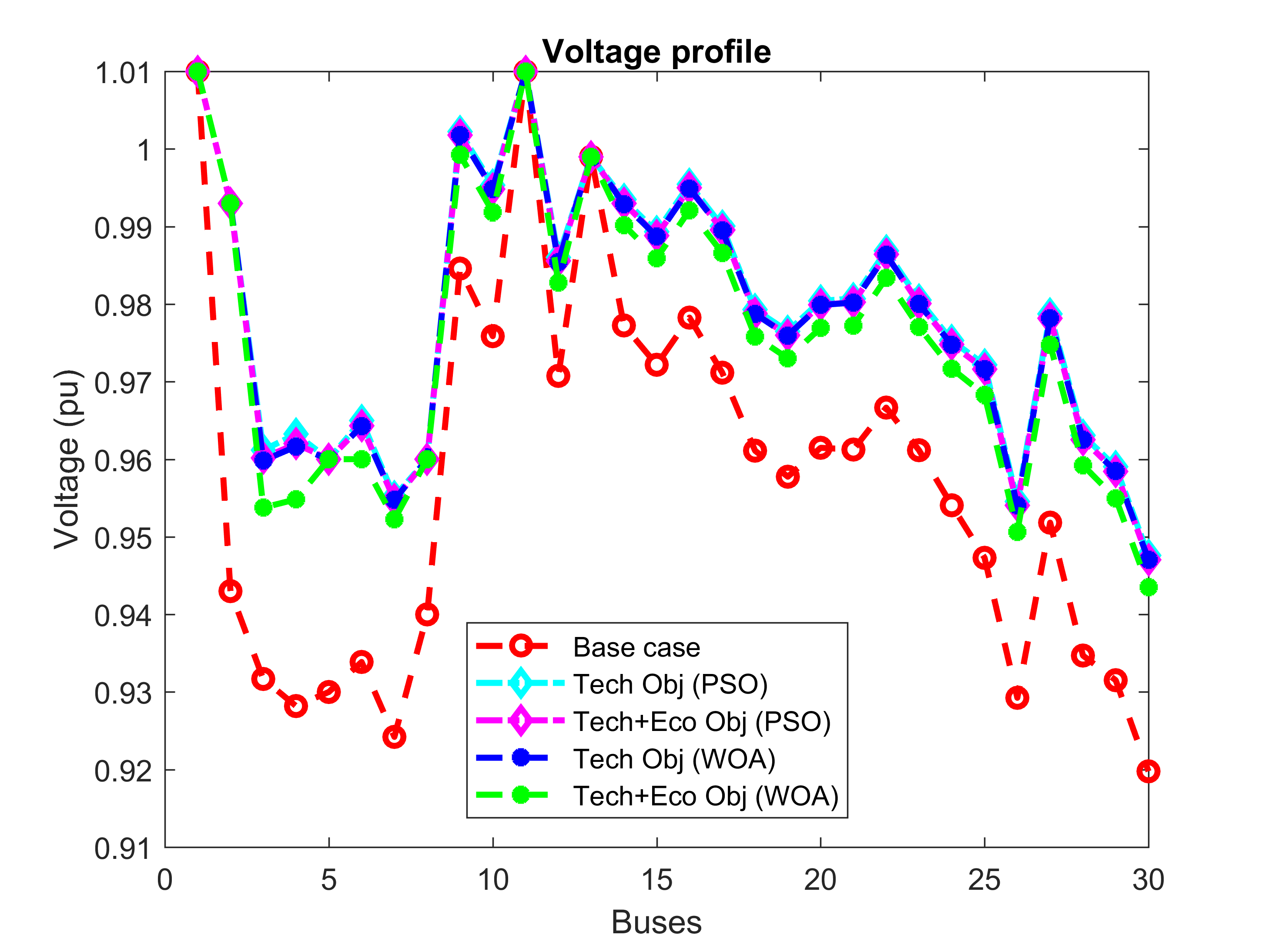}
	\caption{Voltage profile of IEEE 30-bus test system}
	\label{FIG:15}
\end{figure}

\begin{table*}[ht]
\caption{Real and reactive power loss values and improvement for 30-bus test system}\label{tbl4}
\centering
\begin{tabular}{lcc}
\toprule
\textbf{Real power} & \textbf{MW} & \textbf{Performance (\%)} \\
\midrule
WOA Power loss technical & 7.54 & 62.75\% \\
WOA Power loss techno-economic & 10.77 & 132.38\%   \\
PSO Power loss technical & 11.56 & 49.35\% \\
PSO Power loss techno-economic & 11.73 & 51.57\%   \\
\midrule
\textbf{Reactive power} & \textbf{} \\
\midrule
WOA Power loss technical & 34.77 & 50.17\% \\
WOA Power loss techno-economic & 45.87 & 98.15\%   \\
PSO Power loss technical & 48.57 & 40.11\% \\
PSO Power loss techno-economic & 49.17 & 41.82\%   \\
\bottomrule
\end{tabular}
\end{table*}

\begin{table}
\caption{Comparison of the algorithms' real and reactive power loss values for 30-bus test system}\label{tbl4a}
\centering
\begin{tabular}{lcc}
\toprule
\textbf{Real power} & \textbf{Performance (\%)} \\
\midrule

WOA improvement technical & 53.31\%   \\
WOA improvement techno-economic& 8.97\%  \\

\textbf{Reactive power} &  \\
\midrule

WOA improvement technical & 39.71\%   \\
WOA improvement techno-economic & 7.18\%  \\
\bottomrule
\end{tabular}
\end{table}

The LSF was implemented to identify the candidate bus locations for optimal sizing of DGs. The potential bus locations are 2, 6, and 7. These bus locations are selected based on the values close to zero. The first three indices of the bus configuration are selected.

\autoref{FIG:12} illustrates the real power loss on each bus for each algorithm. It is observed that the WOA performs better than the PSO when considering only the technical objective. With the techno-economic objective, the WOA still performs better but by a small margin. \autoref{FIG:13} further shows the summary of results on power loss reduction from the optimization algorithms for the IEEE 30-Bus test system. It is clearly seen that the WOA performs better than the PSO when considering both technical and techno-economic objectives. It is observed from \autoref{tbl4}, that both algorithms decrease the total real power loss. The PSO, under the technical and techno-economic constraints, reduced the real power loss to 11.564 MW and 11.736 MW respectively, while the WOA reduced it to 7.543 MW and 10.770 MW respectively. 

For the optimal sizing, the PSO when considering the technical objective, performs better than the WOA in the sizing of DG units, with the total DG unit's capacity as 58.106 MW compared to the PSOs' 137.547 MW. The WOA has a better total DG unit size when considering the addition of economical objective, with 3 MW lesser than the PSO. However, the PSO produces a lower total DG size, considering the techno-economic objective. However, the WOA produces a lower power loss than the PSO. This is synonymous to WOA results obtained in \cite{Reddy2017a}. The total real and reactive power loss from the PSO is 11.736 MW and 49.167 MVaR, while the WOA is 10.769 MW and 45.870 MVaR respectively. From \autoref{FIG:14}, it is observed that the WOA with techno-economic objective, has a better reactive power loss on the bus network. 

\autoref{FIG:15} shows the voltage profile of the IEEE 30-bus test system. The WOA, with the consideration of techno-economic constraints improved the voltage profile of the bus network.

\begin{table}[h]
\caption{DG units sizes for candidate buses}\label{tbl5}
\centering
\begin{tabular}{lcccc}
\toprule
Bus no. & PSO & PSO (Eco) & WOA & WOA (Eco)\\
\midrule
2 & 8.542 & 33.454 & 45.323 & 23.102 \\
6 & 15.324 & 16.541 & 44.121 & 20.732 \\
7 & 34.240 & 32.330 & 48.103 & 35.442 \\
\bottomrule
\textbf{TOTAL} & \textbf{58.106} & \textbf{82.325} & \textbf{137.547} & \textbf{79.276} \\
\bottomrule
\end{tabular}
\end{table}

\section{Conclusion}\label{section7}
In this paper, swarm intelligence-based algorithms, the PSO and the WOA were applied to the sizing of DG units for optimal planning of a meshed transmission network. The paper reports a proper evaluation of the application of both algorithms on the IEEE bus test systems (14- and 30-bus). Technical and techno-economic (addition of both technical and economical) objectives were considered, and were simulated separately for more observations. 

The algorithms outperform each other in different metrics. For instance, the WOA when considering the techno-economic objective, performs better in the total real power loss minimization for both the IEEE 14- and 30-bus test systems. The WOA also performs better in the voltage profile improvement of the IEEE 14-bus network while the PSO ranks higher in performance on the voltage profile improvement for the IEEE 30-bus network. However, the PSO performs better when considering only the technical objective for the total power loss minimization. The reactive power loss is most minimized by the WOA when considering the techno-economic objective for the IEEE 14-bus network. However, the WOA performs better on the 30-bus network, when the technical objective is only considered.

The PSO performs better in the sizing of the DG units, when considering the technical and techno-economic objectives for the IEEE 14- and 30-bus network. The only exception (where the WOA performs better) is the consideration of the technical objective for the IEEE 14-bus network.

The consideration of the technical and techno-economic objectives have their differences. The technical objective can be more generalized since it is more scientific while the addition of the economical objective (which is techno-economic) makes it less generalizable. This is because economic parameters are dependent on a country's economic factors. In some cases, the economic state of provinces or states in a country may vary. Furthermore, economic factors are also time-dependent. 


Results also proved that the PSO and the WOA remain successful in optimal placement and sizing problem even in transmission networks. Both algorithms produced a remarkable result from the solutions. PSO produces a lesser total DG units size from the IEEE 14-bus test system evaluation. The WOA had the most minimal real and reactive power loss values on both bus configuration. The WOA also performs better but by a few margin in the voltage profile improvement. 

For future studies, newly modified or hybridized algorithms can be developed according to the specific lapses in present results.

\bibliographystyle{unsrt}      
\bibliography{references11}   

\end{document}